\documentclass{article}

\usepackage{microtype}
\usepackage{graphicx}
\usepackage{subfigure}
\usepackage{booktabs} 
\usepackage{amsmath}%
\usepackage{amsfonts}
\usepackage{multirow}
\usepackage{mathtools}
\usepackage{ marvosym }
\usepackage{tikz}
\usetikzlibrary{bayesnet}

\usepackage[accepted]{icml2018}
\newcommand{\R}{\mathcal{R}}
\newcommand{\vnorm}[1]{\left|\left|#1\right|\right|}
\newcommand{\bigCI}{\mathrel{\text{\scalebox{1.07}{$\perp\mkern-10mu\perp$}}}}
\icmltitlerunning{Black Box FDR}

\begin{document}

\twocolumn[
\icmltitle{Black Box FDR}



\icmlsetsymbol{equal}{*}

\begin{icmlauthorlist}
\icmlauthor{Wesley Tansey}{dsi,cumc}
\icmlauthor{Yixin Wang}{dsi,stats}
\icmlauthor{David M.~Blei}{dsi,stats,cs}
\icmlauthor{Raul Rabadan}{cumc}
\end{icmlauthorlist}

\icmlaffiliation{dsi}{Data Science Institute, Columbia University, New York, NY, USA}
\icmlaffiliation{cumc}{Department of Systems Biology, Columbia University Medical Center, New York, NY, USA}
\icmlaffiliation{stats}{Department of Statistics, Columbia University, New York, NY, USA}
\icmlaffiliation{cs}{Department of Computer Science, Columbia University, New York, NY, USA}

\icmlcorrespondingauthor{Wesley Tansey}{wt2274@columbia.edu}

\icmlkeywords{Hypothesis Testing, False Discovery Rate, FDR, Deep Learning, Feature Selection}

\vskip 0.3in
]



\printAffiliationsAndNotice{}  

\begin{abstract}
Analyzing large-scale, multi-experiment studies requires scientists to test each experimental outcome for statistical significance and then assess the results as a whole. We present Black Box FDR (BB-FDR), an empirical-Bayes method for analyzing multi-experiment studies when many covariates are gathered per experiment. BB-FDR learns a series of black box predictive models to boost power and control the false discovery rate (FDR) at two stages of study analysis. In Stage 1, it uses a deep neural network prior to report which experiments yielded significant outcomes. In Stage 2, a separate black box model of each covariate is used to select features that have significant predictive power across all experiments. In benchmarks, BB-FDR outperforms competing state-of-the-art methods in both stages of analysis. We apply BB-FDR to two real studies on cancer drug efficacy. For both studies, BB-FDR increases the proportion of significant outcomes discovered and selects variables that reveal key genomic drivers of drug sensitivity and resistance in cancer.


\end{abstract}


\section{Introduction}
\label{sec:introduction}
High-throughput screening (HTS) techniques have fundamentally changed the landscape of modern biological experimentation. Rather than conducting just one experiment at a time, HTS enables scientists to perform hundreds of parallel experiments, each with different biological samples and different interventions. At the same time, HTS also enables scientists to gather rich contextual information about each experiment by profiling the samples under study using techniques like DNA sequencing. Thus, each HTS study produces a dataset of many experiments, where each experiment contains both an outcome variable and a high-dimensional feature set describing the context.

\begin{figure*}[th]
\centering
\includegraphics[width=\textwidth]{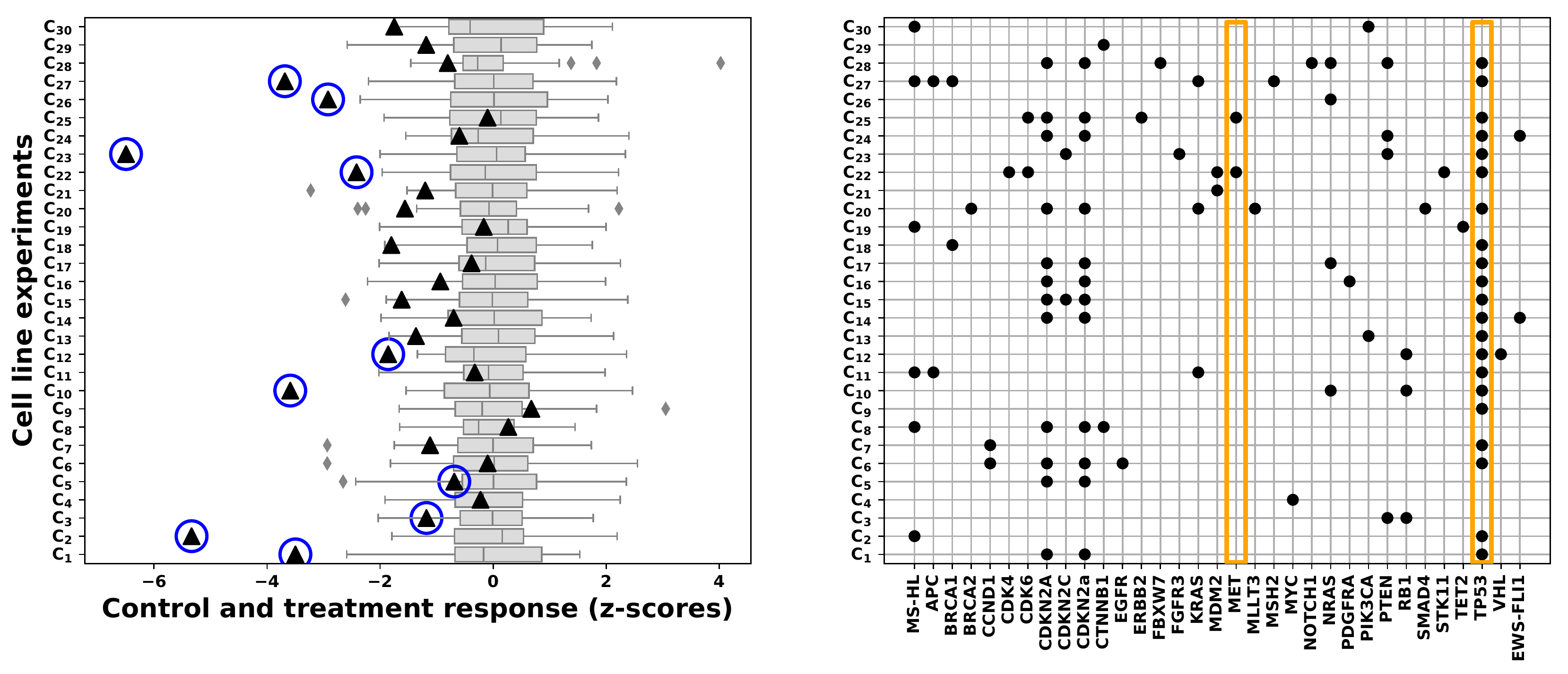}
\caption{\label{fig:study_example} Left: a subset of $30$ cell line experiments from the Nutlin-3 case study in Section~\ref{sec:cancer}. Control replicates (gray box plots) and cell line responses (black triangles) are measured as $z$-scores relative to mean control values. Right: a subset of the corresponding genomic features for each experiment; black dots indicate a cell line has a recurrent mutation in the gene labeled on the x-axis. The goal in Stage 1 analysis is to select cell lines that showed a significant response (circled in blue). In Stage 2, the genomic features are analyzed to understand the mutations driving drug response (circled in orange).}
\end{figure*} 

Figure~\ref{fig:study_example} shows a slice of the Genomics of Drug Sensitivity in Cancer (GDSC) dataset \cite{yang:etal:2012:gdsc}, an HTS study investigating how cancer cell lines respond to different cancer therapeutics. The left panel shows the relative response of 30 different cancer cell lines $(C_1, C_2, \ldots, C_{30})$ treated with the drug Nutlin-3. For each cell line, the treatment response (black triangles) is overlayed on top of the untreated control replicate distribution (gray box plots). Even when no drug is applied, each cell line still exhibits natural variation. The first goal in analyzing this data is therefore to address the question of whether a given cell line responded to the treatment. Concretely, we need to perform a hypothesis test for each cell line, where the null hypothesis is that the drug had no effect. Absent other information, this would be a classic \textit{multiple hypothesis testing} (MHT) problem. 

But HTS studies such as GDSC differ from the classical setup by also producing a rich set of side information for each experiment. The right panel of Figure~\ref{fig:study_example} shows a subset of the genomic profile for each cell line, with a black dot indicating the cell line has a mutation in that gene. Biologically, a mutated gene can lead to different phenotypic behavior that may cause sensitivity or resistance to a drug.

Statistically, this means the likelihood of a cell line responding to treatment is a latent function of that cell line's genomic profile. Identifying which mutations are associated with treatment response could guide future experiments and development of new targeted therapies. Deriving scientific insight from patterns across experiments represents a second phase of hypothesis testing, where the null hypothesis is that a given gene is not associated with drug response.

We term these two phases \textit{Stage 1} and \textit{Stage 2} and ask two scientifically-motivated statistical inference questions:
\begin{itemize}
\item \textbf{Stage 1}: How do we leverage the available side information in a HTS study to increase how many significant outcomes we can detect?
\item \textbf{Stage 2}: Can we discover which variables are associated with significant outcomes, even when the underlying function is high-dimensional and nonlinear?
\end{itemize}

We answer both of these questions and propose Black Box FDR (BB-FDR), a method for analyzing multi-experiment studies with many covariates gathered per experiment. BB-FDR uses the covariates to build a deep probabilistic model that predicts how likely a given experiment is to generate significant outcomes \textit{a priori}. It uses this prior model to adaptively select significant outcomes in a manner that controls the overall false discovery rate (FDR) at a specified Stage 1 level. BB-FDR then builds a predictive model of each covariate to perform variable selection on the Stage 1 model while conserving a specified Stage 2 FDR threshold.

We validate BB-FDR on both synthetic and real data. BB-FDR outperforms other state-of-the-art Stage 1 methods in a series of benchmarks, including the recently-proposed NeuralFDR~\cite{xia:etal:2017:neuralfdr}. BB-FDR is also a more pragmatic choice compared to a fully-Bayesian approach: it scales trivially to thousands of covariates, can learn arbitrarily complex functions, and runs easily on a laptop. We apply BB-FDR to a real-world case study of two cancer drug screenings. BB-FDR finds more significant discoveries on the data and recovers an informative set of biologically-plausible genes that may convey drug sensitivity and resistance in cancer.


\section{Multiple testing and FDR control}
\label{sec:background}
In the classical MHT setup, $\mathbf{z} = (z_1, \ldots, z_n)$ are a set of independent observations of the outcomes of $n$ experiments. For each experiment, a treatment is applied to a target and the treatment has either no effect ($h_i = 0$) or some effect ($h_i = 1$). If the treatment has no effect, the distribution of the test statistic is the null distribution $f_0(z)$; otherwise it follows an unknown alternative distribution $f_1(z)$. The null hypothesis for every experiment is that the test statistic was drawn from the null distribution: $H_0: h_i = 0$.

\subsection{False discovery rate control}
\label{subsec:background:fdr}
In most experiments of interest, it is impossible to determine $h_i$ with no error. For a given prediction $\hat{h}_i$, we say it is a true positive or a true discovery if $\hat{h}_i = 1 = h_i$ and a false positive or false discovery if $\hat{h}_i = 1 \neq h_i$. Let $\mathcal{S} = \{i : h_i = 1\}$ be the set of observations for which the treatment had an effect and $\hat{\mathcal{S}} = \{i : \hat{h}_i = 1\}$ be the set of predicted discoveries. We seek procedures that maximize the true positive rate (TPR) also known as \textit{power}, while controlling the false discovery rate--the expected proportion of the predicted discoveries that are actually false positives,
\begin{equation}
\label{eqn:fdp}
\text{FDR} \coloneqq \mathbb{E}[\text{FDP}] \, , \quad \quad \text{FDP} = \frac{\#\{ i : i \in \hat{\mathcal{S}} \backslash \mathcal{S} \}}{\#\{ i : i \in \hat{\mathcal{S}} \}} \, .
\end{equation}
FDP in \eqref{eqn:fdp} is the \textit{false discovery proportion}: the actual proportion of false positives in the predicted discovery set for a specific experiment. While ideally we would like to control the FDP, the randomness of the outcome variables makes this impossible in practice. Thus FDR is the typical error measure targeted in modern scientific analyses.

\subsection{Related work}
Controlling FDR in multiple hypothesis testing has a long history in statistics and machine learning. The Benjamini-Hochberg (BH) procedure \cite{benjamini:hochberg:1995:bh} is the classic technique and still the most widely used in science. Many other methods have since been developed to take advantage of study-specific information to increase power. Recent examples include accumulation tests for ordering information \cite{li:barber:2017:accumulation}, the p-filter for grouping and test statistic dependency \cite{ramdas:etal:2017:pfilter}, FDR smoothing for spatial testing \cite{tansey:etal:2017:fdrsmoothing}, FDR-regression for low-dimensional covariates \cite{scott:etal:2015:fdrreg}, and, most recently, NeuralFDR for high-dimensional covariates \cite{xia:etal:2017:neuralfdr}. We consider high-dimensional covariates and compare against NeuralFDR in Section~\ref{sec:benchmarks}.


\section{Black Box FDR}
\label{sec:method}
Consider a study with $n$ independent experiments that produces a set of independent test statistics $\mathbf{z} = (z_1, \ldots, z_n)$ corresponding to the outcome measurements, as in Section~\ref{sec:background}. However, now each experiment also has a vector of $m$ covariates $X_{i\cdot} = (X_{i1}, \ldots, X_{im})$ containing side information that may influence the outcome of that experiment. Specifically, whether the experiment comes from the null distribution $h_i=0$ or the alternative $h_i=1$ is allowed to depend arbitrarily on $X_{i\cdot}$.

BB-FDR extends the empirical-Bayes two-groups model of \citet{efron:2008:twogroups} by building a hierarchical probabilistic model with experiment-specific priors modeled through a deep neural network. We first estimate the alternative distribution offline using predictive recursion \cite{newton:2002} to estimate $f_1$. This follows other recent extensions to the two-groups model \cite{scott:etal:2015:fdrreg,tansey:etal:2017:fdrsmoothing} and enjoys strong empirical performance and consistency guarantees \cite{tokdar:martin:ghosh:2009}. BB-FDR then focuses on modeling the experiment-specific prior, assuming the null and alternative distributions are fixed.



\begin{figure}[t]
\centering
\begin{tikzpicture}

  \node[obs]                               (z) {$z$};
  \node[latent, above=of z]                (h) {$h$};
  \node[latent, right=of h]                (c) {$c$};
  \node[latent, right=of c]                (a) {$a$};
  \node[latent, below=of a]                (b) {$b$};
  \node[obs, right=of a, yshift=-0.8cm]                (x) {$x$};
  \node[det, above=of x]                (t) {$\theta$};

  \edge {h} {z} ; %
  \edge {c} {h}  ; %
  \edge {a, b} {c}  ; %
  \edge {x} {a, b} ;
  \edge {t} {a, b} ;

  \plate {zx} {(z)(h)(c)(a)(b)(x)} {$n$} ;
  \plate {x} {(x)} {$M$} ;

\end{tikzpicture}
\caption{\label{fig:graphical_model} The graphical model for BB-FDR.}
\end{figure}
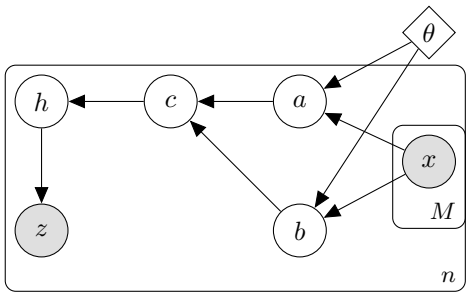

\subsection{Stage 1: determining significant outcomes}
\label{subsec:method:twogroups}
We model the test statistic as arising from a mixture model of two components, the null ($f_0$) and the alternative ($f_1$). An experiment-specific weight $c_i$ then models the prior probability of the test statistic coming from the alternative (i.e. the probability of the treatment having an effect \textit{a priori}). We place a beta prior on each experiment-specific prior $c_i$ and model the parameters of the hyperprior with a black box function $G$ parameterized by $\theta$; in our implementation, $G$ is a deep neural network. The complete model for BB-FDR is:
\begin{equation}
\label{eqn:bbfdr_generative_model}
\begin{aligned}
z_i &\sim h_i f_1(z_i) + (1-h_i) f_0(z_i) \\
h_i &\sim \text{Bernoulli}(c_i) \\
c_i &\sim \text{Beta}(a_i, b_i) \\
a_i, b_i &= G_{\theta,i}(X) \, .
\end{aligned}
\end{equation}
We optimize $\theta$ by integrating out $h_i$ and maximizing the complete data log-likelihood,
\begin{equation}
\label{eqn:bbfdr_data_likelihood}
p_\theta(z_i) = \int_0^1 (c_i f_1(z_i) + (1-c_i) f_0(z_i)) \text{Beta}(c_i | G_{\theta,i}(X)) dc_i \, .
\end{equation}
Figure~\ref{fig:graphical_model} shows the BB-FDR graphical model.

The beta prior is a departure from other two-groups extensions, which use a flatter hierarchy and learn a predictive model for $c_i$ \cite{scott:etal:2015:fdrreg,tansey:etal:2017:fdrsmoothing}. We found the flat approach to be difficult to train, leading to a degenerate $G$ that always predicts the global mean prior. 

A hierarchical prior improves training for two reasons. First, optimization is easier and more stable because the output of the function is two soft-plus activations. Compared to a sigmoid, this form leads to less saturated gradients. Second, the additional hierarchy allows the model to assign different degrees of confidence to each experiment, changing the model from homoskedastic to heteroskedastic. Finally, we found it important to enforce concavity of the beta distribution; we thus add $1$ to both $a_i$ and $b_i$.

We fit the model in \eqref{eqn:bbfdr_generative_model} with stochastic gradient descent (SGD) on an $L_2$-regularized loss function,
\begin{equation}
\label{eqn:bbfdr_objective}
\begin{aligned}
& \underset{\theta \in \R^{|\theta|}}{\text{minimize}}
& & 
-\sum_{i} \log
p_\theta(z_i) + \lambda \vnorm{G_\theta(X)}_F^2 \, ,
\end{aligned}
\end{equation}
where $\vnorm{\cdot}_F$ is the Frobenius norm. In pilot studies, we found adding a small amount of $L_2$-regularization prevented over-fitting at virtually no cost to statistical power. For computational purposes, we approximate the integral in \eqref{eqn:bbfdr_data_likelihood} by a fine-grained numerical grid. 


Once the optimized parameters $\hat{\theta}$ are chosen, we calculate the posterior probability of each test statistic coming from the alternative,
\begin{flalign}
\label{eqn:bbfdr_posterior}
\hat{w}_i &= p_{\hat{\theta}}(h_i = 1 | z_i) \\\nonumber
&= \int_0^1 \frac{c_i f_1(z_i) \text{Beta}(c_i | G_{\hat{\theta},i}(X))}{c_i f_1(z_i) + (1-c_i) f_0(z_i)} dc_i \, .
\end{flalign}
Assuming the posteriors are accurate, rejecting the $i^{\text{th}}$ hypothesis will produce $1-\hat{w}_i$ false positives in expectation. Therefore we can maximize the total number of discoveries by a simple step down procedure. First, we sort the posteriors in descending order by the likelihood of the test statistics being drawn from the alternative. We then reject the first $q$ hypotheses, where $0 \leq q \leq n$ is the largest possible index such that the expected proportion of false discoveries is below the FDR threshold. Formally, this procedure solves the optimization problem,
\begin{equation}
\label{eqn:step_down_procedure}
\begin{aligned}
& \underset{q}{\text{maximize}}
& & 
q \\
& \text{subject to} & & \frac{\sum_{i=1}^q (1-\hat{w}_i)}{q} \leq \alpha \, ,
\end{aligned}
\end{equation}
for a given FDR threshold $\alpha$. By convention $\frac{0}{0} = 0$.

The model in \eqref{eqn:bbfdr_generative_model}--\eqref{eqn:step_down_procedure} handles Stage 1 of the analysis. The black box model $G$ uses the entire feature vector $X_{i\cdot}$ of every experiment to predict the prior parameters over $c_i$. The observations $z_i$ are then used to calculate the posterior probability $\hat{w}_i$ that the treatment had an effect. The selection procedure in \eqref{eqn:step_down_procedure} uses these posteriors to reject a maximum number of null hypotheses while conserving the FDR.

\subsection{Stage 2: identifying important variables}
\label{subsec:method:covariates}
Using a flexible black box model for $G$ in \eqref{eqn:bbfdr_generative_model} provides a trade-off. On one hand, it enables BB-FDR to learn a rich class of functions for the relationship between the covariates and the test statistic. As we show in Section~\ref{sec:benchmarks}, this increases power in Stage 1 compared to a standard linear model.

However, variable selection (Stage 2) is straightforward in linear models whereas black box models are by definition opaque. Understanding which variables deep learning models use to make predictions is an ongoing area of research in both machine learning \citep[e.g.][]{elenberg:etal:2017:interpretation} and specific scientific disciplines \citep[e.g.][in ecology]{olden:jackson:2002:ann_variables,giam:olden:2015:ann_variables}. As far as we are aware, there are currently no methods that provide variable selection with FDR control when the covariates may have arbitrary dependency structure.


To overcome this challenge, BB-FDR uses conditional randomization tests (CRTs) \cite{candes:etal:2018:panning}. The idea of a CRT is to model each coordinate of the feature matrix $X_{\cdot j}$ using only the other coordinates $X_{\cdot -j}$. The conditional distribution $\mathbb{P}(X_{\cdot j} | X_{\cdot -j})$ then represents a valid null distribution for testing the hypothesis $X_{\cdot j} \bigCI Z | X_{\cdot -j}$, where $Z$ is the test statistic. The corresponding $p$-value can be calculated by sampling from the conditional to approximate the true $p$-value,
\begin{equation*}
\label{eqn:crt_p_value}
p_{j} = \mathbb{E}_{\widetilde{X}_{\cdot j} \sim \mathbb{P}(X_{\cdot j} | X_{\cdot -j})}\left[ \mathbb{I}\left[ t(\mathbf{z}, X) \leq t(\mathbf{z}, (\widetilde{X}_{\cdot j}, X_{\cdot -j})) \right] \right] \, ,
\end{equation*}
where $t$ is the test statistic of interest. Once the $p$-values have been estimated for all features, we can apply standard BH correction and report significant features.

BB-FDR tests which features are associated with a change in the posterior probability of $z_i$ coming from the alternative. It uses the negative entropy of the posteriors as the test statistic,
\begin{equation}
\label{eqn:crt_t_stat}
t(\mathbf{z}, X) = \sum_i \hat{w}_i \log \hat{w}_i + \sum_i (1-\hat{w}_i) \log (1-\hat{w}_i) \, .
\end{equation}
Intuitively, if a feature is useful in predicting treatment efficacy, it should reduce the overall entropy of the posterior. By definition, a feature sampled from the null adds no new information to the model; it cannot systematically reduce the entropy.

For this procedure to retain frequentist consistency guarantees, both the conditional null distribution and the model of the prior must be the true distributions. In practice, one never has access to these and thus we estimate both; for the conditional null, we use gradient boosting trees \cite{chen:guestrin:2016:xgboost}. 

\section{Benchmarks}
\label{sec:benchmarks}
We perform a series of benchmark studies to assess the performance of BB-FDR in both stages of inference. For each benchmark, we compare the power of BB-FDR to other state-of-the-art approaches. In all studies, we consider binary covariates and real-valued z-scores as test statistics.

Across experiments, we found BB-FDR is particularly suitable for large samples: it outperforms competing methods in both stages while being more computationally efficient.

\begin{figure}
\centering
\includegraphics[width=0.45\textwidth]{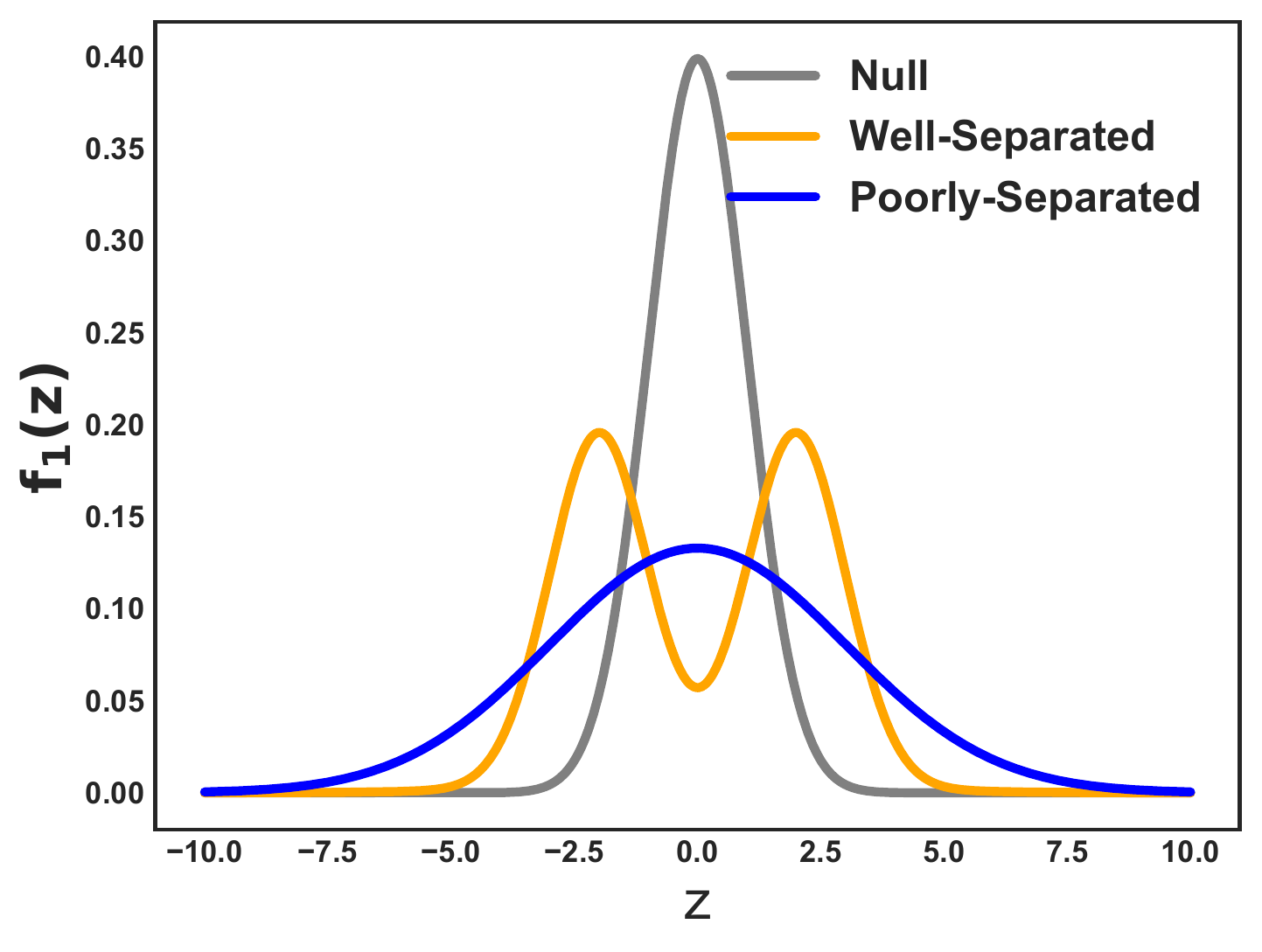}
\caption{The two alternative densities used in our benchmarks. The well-separated (WS) density has little overlap with the null, making for a stronger signal.}
\label{fig:alternative_densities}
\end{figure}

\begin{figure*}[p]
\centering
\subfigure[Constant (PS)]{\includegraphics[width=0.45\textwidth]{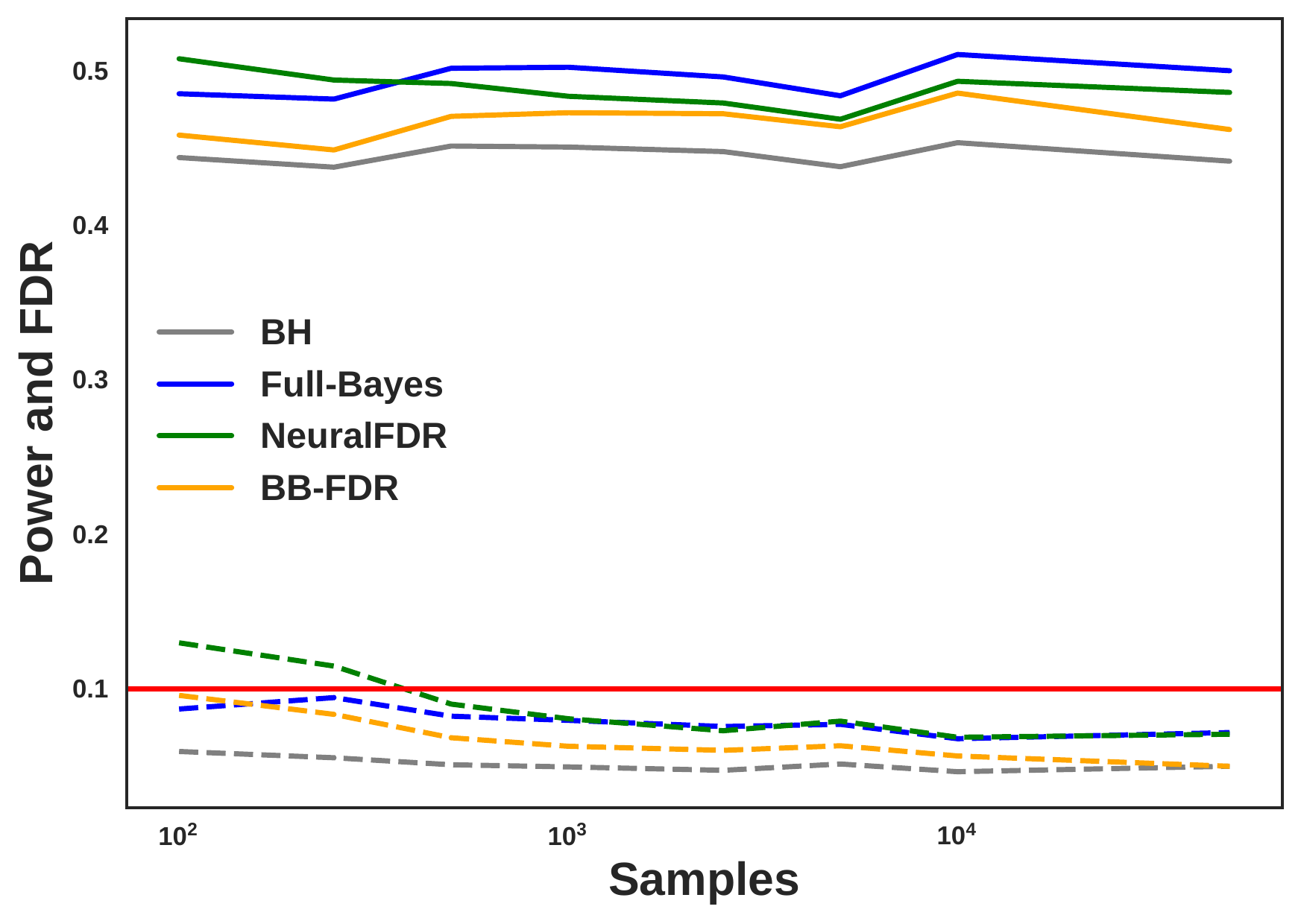}}
\subfigure[Constant (WS)]{\includegraphics[width=0.45\textwidth]{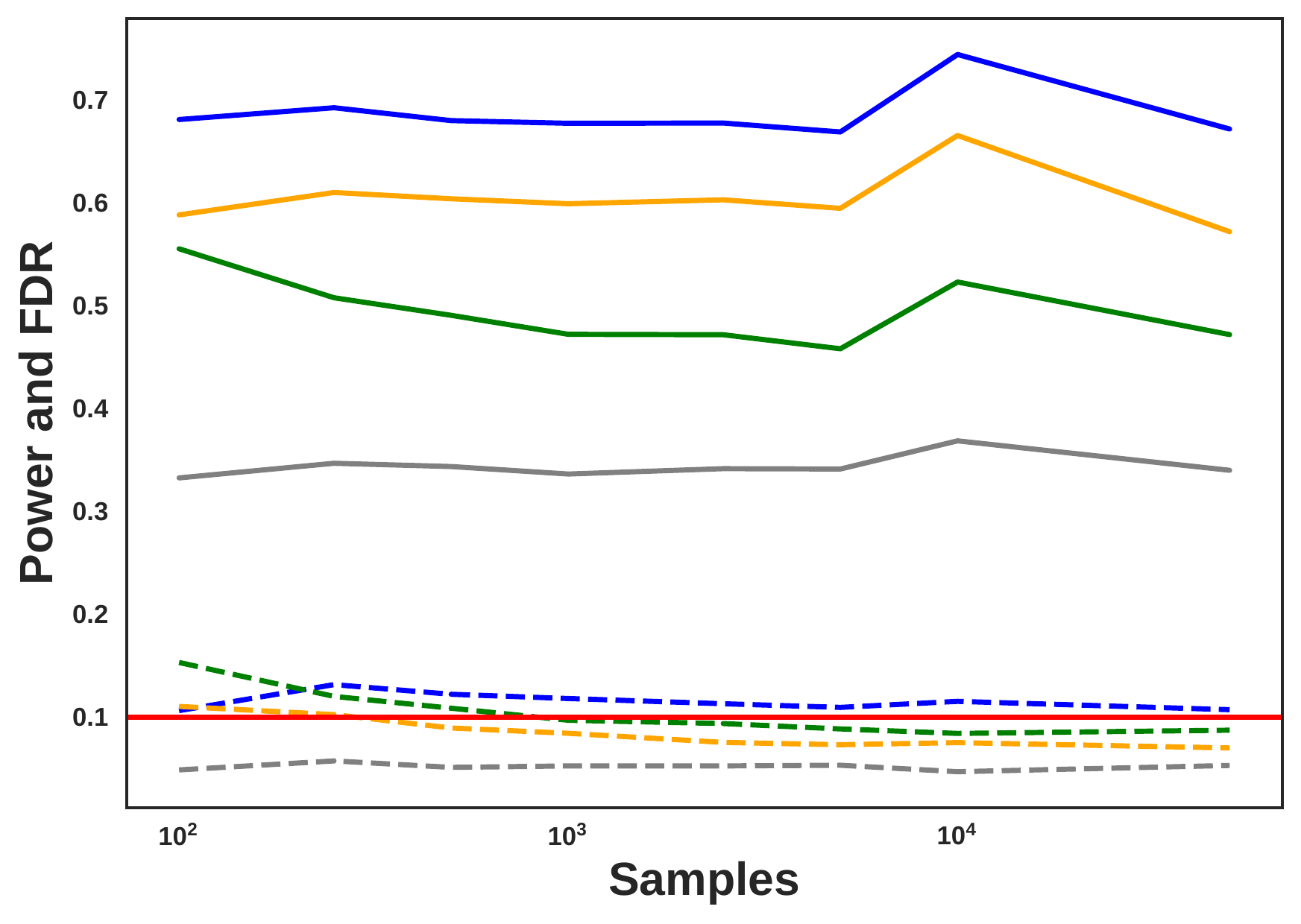}}
\subfigure[Linear (PS)]{\includegraphics[width=0.45\textwidth]{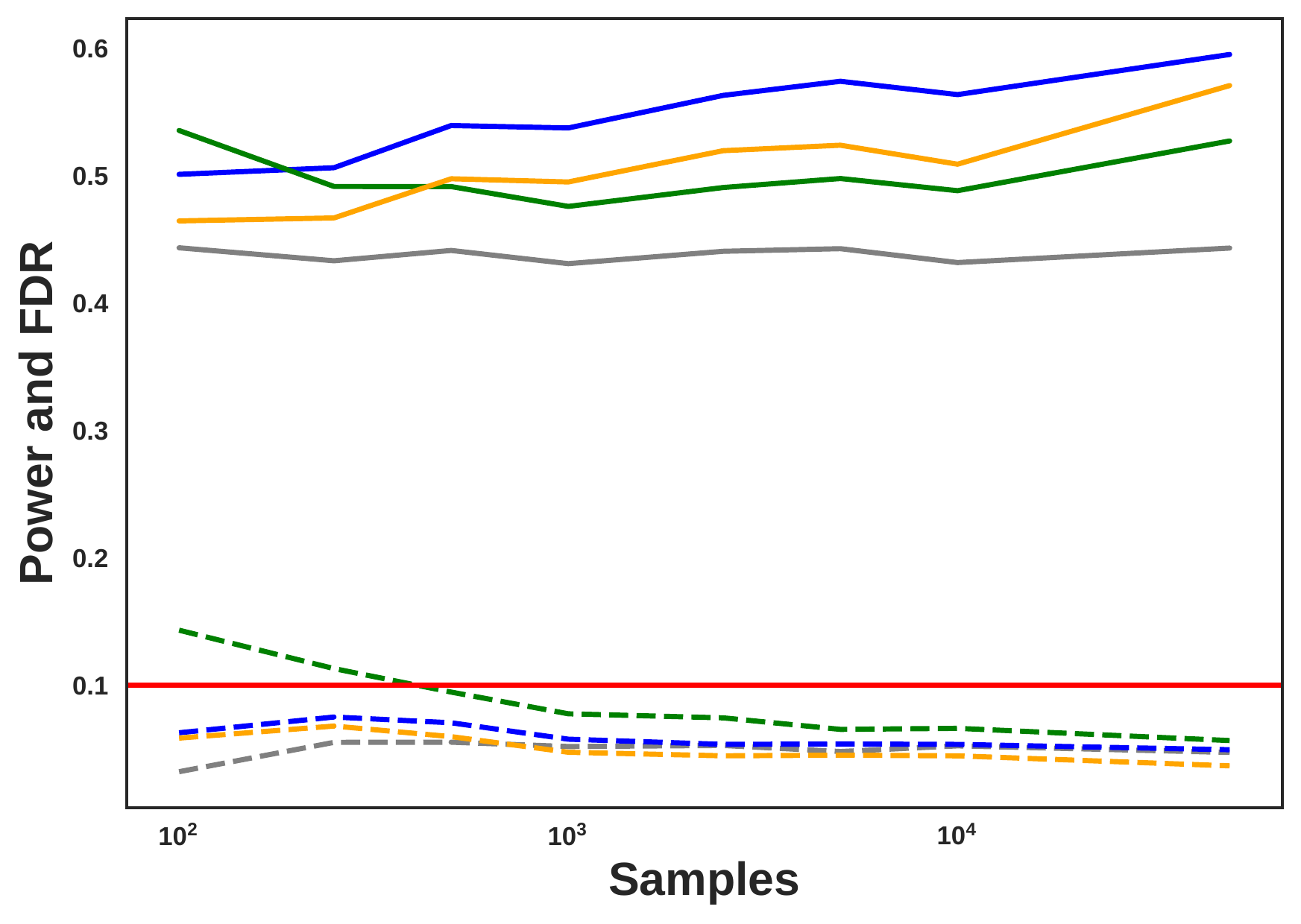}}
\subfigure[Linear (WS)]{\includegraphics[width=0.45\textwidth]{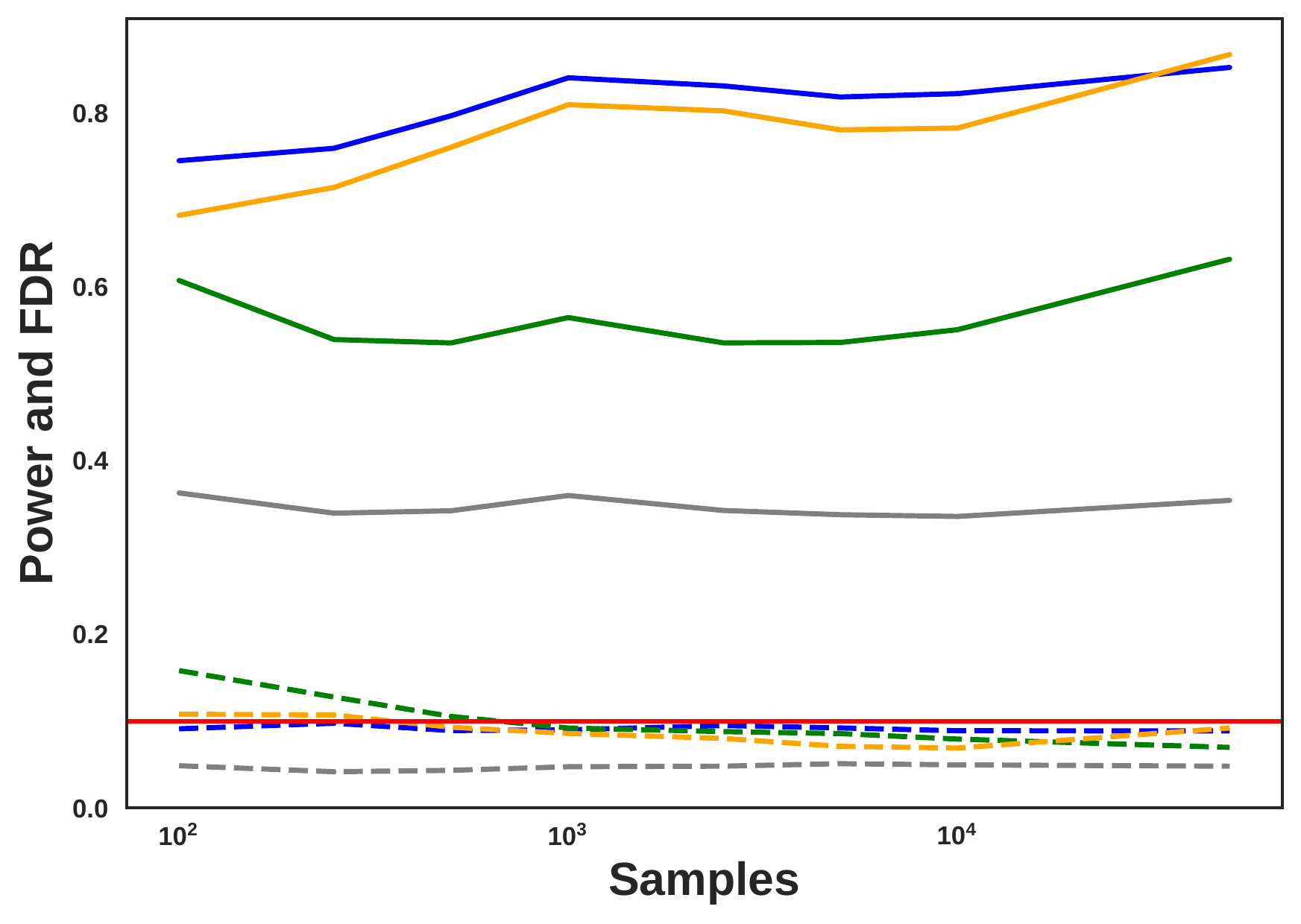}}
\subfigure[Nonlinear (PS)]{\includegraphics[width=0.45\textwidth]{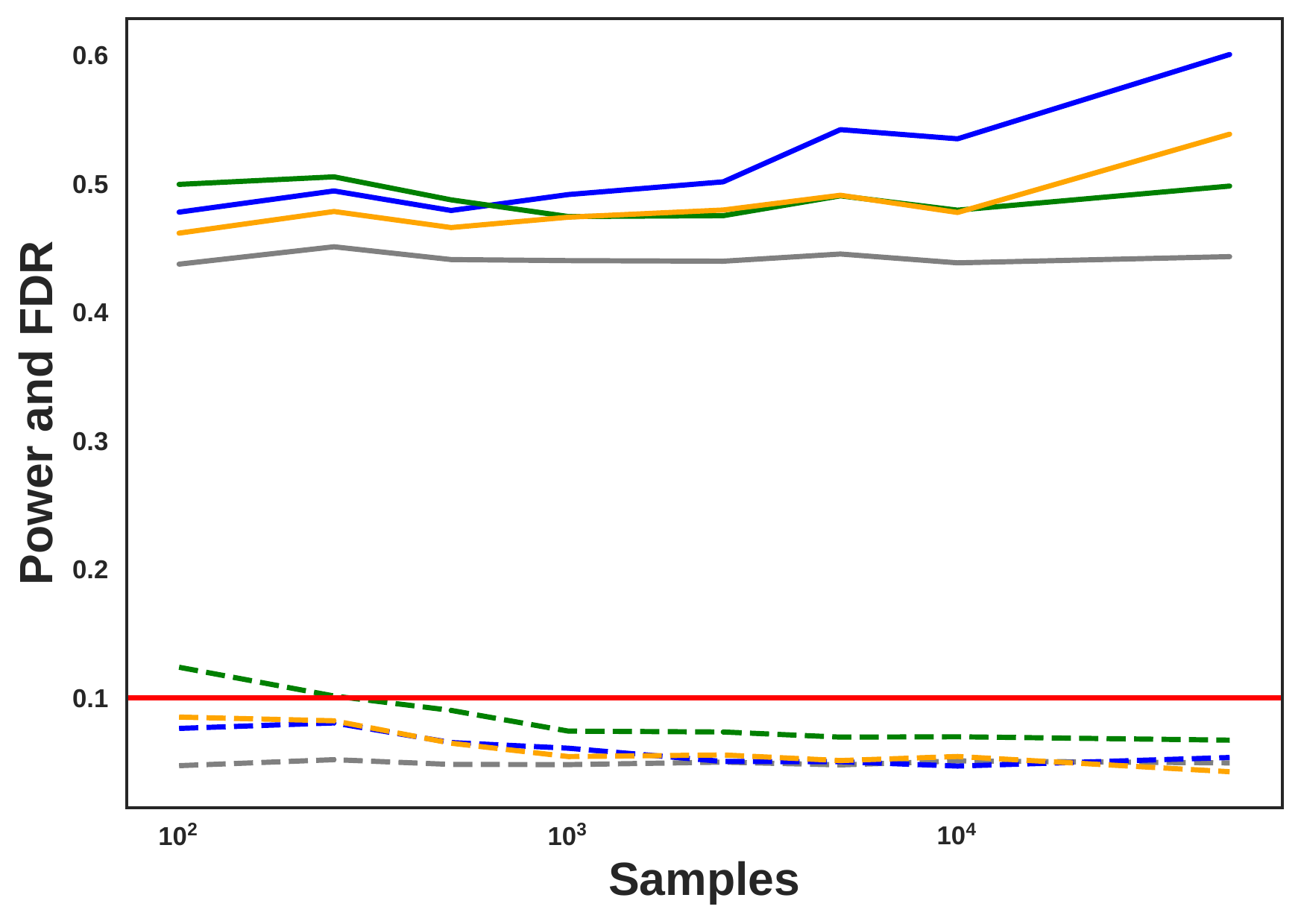}}
\subfigure[Nonlinear (WS)]{\includegraphics[width=0.45\textwidth]{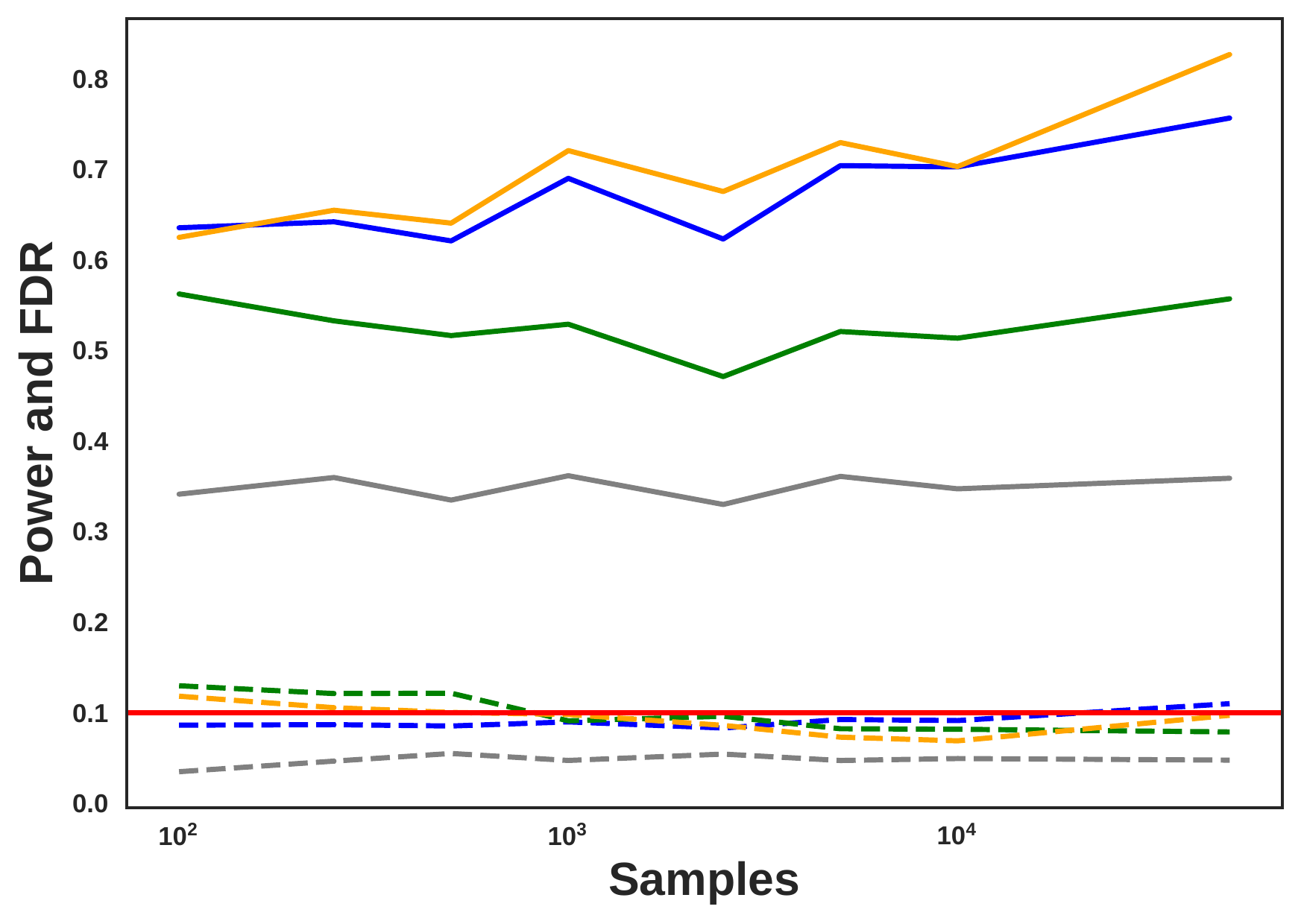}}
\caption{\label{fig:benchmarks} Hypothesis testing results on the synthetic datasets averaged over 100 trials at varying sample sizes on the two different alternative distributions. Solid lines show power; dashed lines show estimated FDR; the red horizontal line denotes the specified $10\%$ FDR threshold. In general, the Benjamini-Hochberg and NeuralFDR methods have lower power since they do not model the alternative. The fully-Bayesian method has high power in the low-to-moderate sample regime, but as the sample size grows the empirical-Bayes approach of BB-FDR becomes more effective.}
\end{figure*}

\begin{figure*}[ptbh]
\centering
\subfigure[Linear (PS)]{\includegraphics[width=0.45\textwidth]{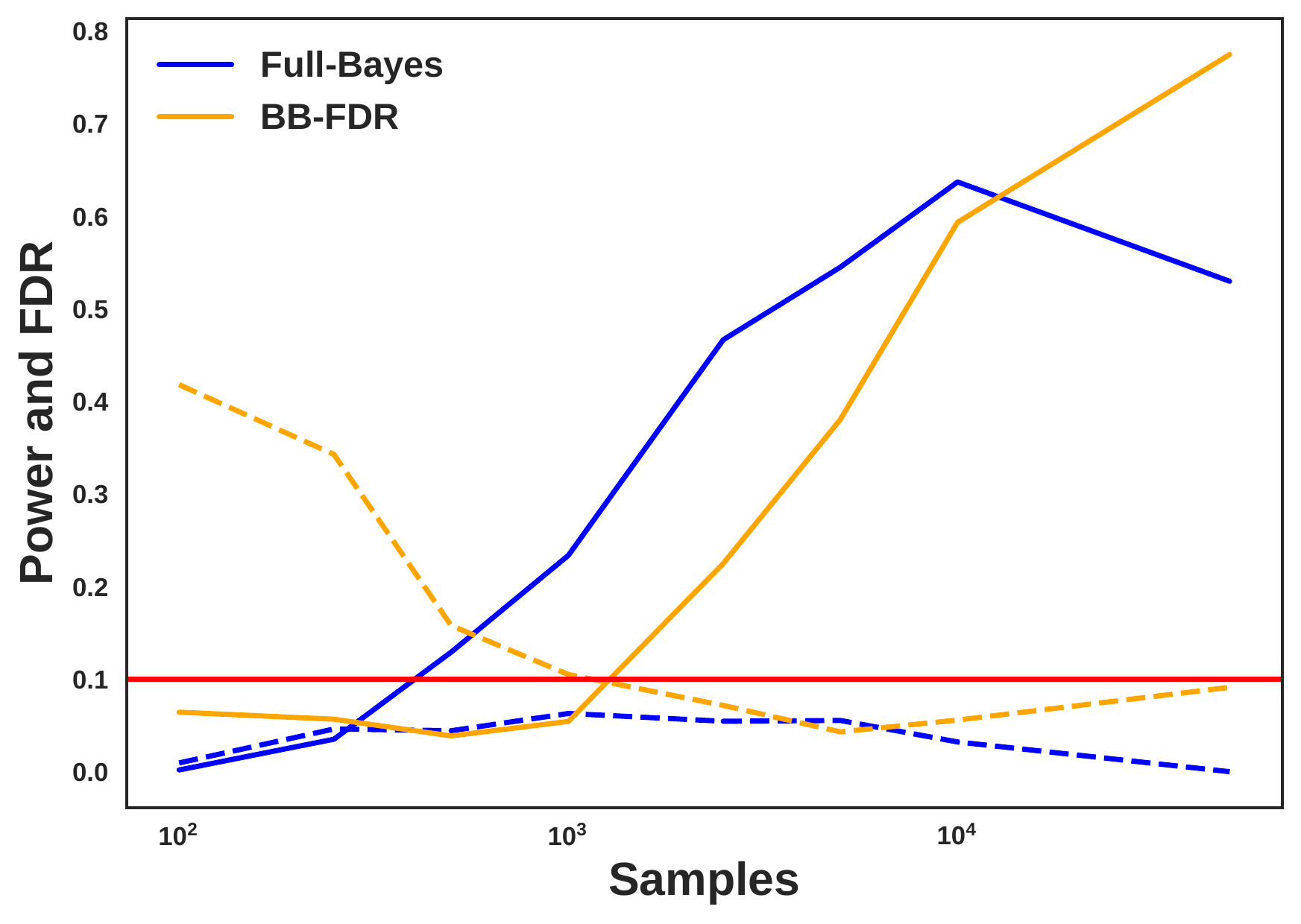}}
\subfigure[Nonlinear (PS)]{\includegraphics[width=0.45\textwidth]{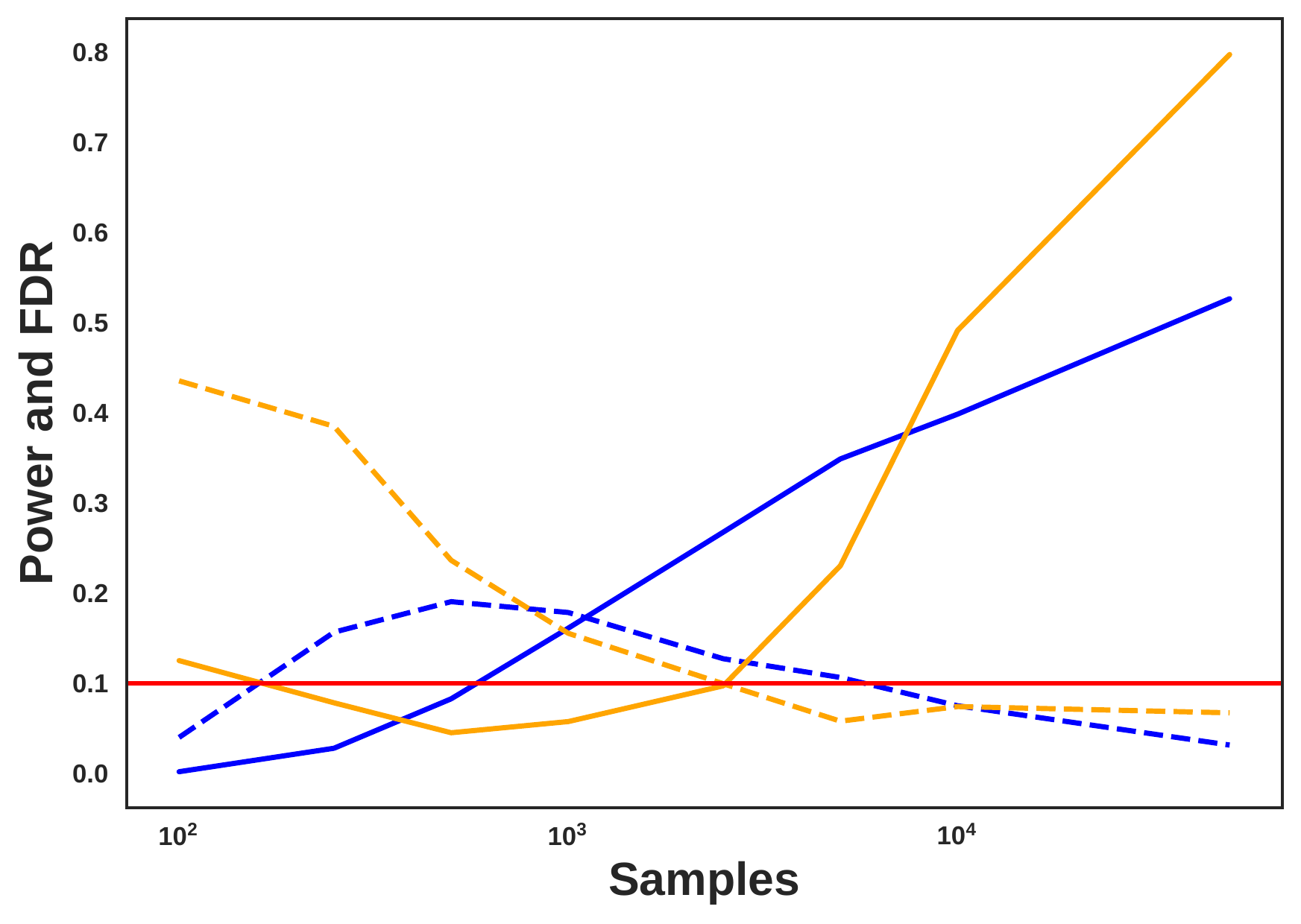}}
\caption{\label{fig:benchmarks_covariates} Variable selection results at a 10\% FDR threshold. In low sample regimes, the conditional null distribution used in the CRT procedure is poorly fit and results in violated FDR thresholds. At moderate-to-large samples, BB-FDR has higher power than the fully-Bayesian model and conserves FDR.}
\end{figure*}

\subsection{Setup}
\label{subsec:benchmarks:setup}
We consider three different ground truth models for $P(X)$, the joint distribution over the covariates, and $P(h=1 | X)$, the prior probability of coming from the alternative distribution given the covariates:
\begin{itemize}
    \item \textbf{Constant}: All covariates are sampled IID normal; the prior is independent of the covariates, with $P(h_i=1 | X) = 0.5$.
    \item \textbf{Linear}: Covariates are sampled from a multivariate normal with full covariance matrix (i.e. conditionally linear); the prior is a linear function with IID standard normal coefficients for each covariate.
    \item \textbf{Nonlinear}: Covariates and prior coefficients are generated similarly. We first drawing 20 IID uniform $\text{Bern}(0.5)$ latent variables. For each covariate, 5 pairs of latent variables $(u_i, u_j)$ are chosen and with equal probability are either \texttt{AND}ed or \texttt{XOR}ed together and multiplied by a draw from a standard normal; the latent weights are summed to get the final logit value for the covariate or coefficient.
\end{itemize}
For each of the three ground truth models, we consider two different alternative distributions:
\begin{itemize}
    \item \textbf{Well-Separated (WS)}: A 3-component Gaussian mixture model, $f_1(z) = 0.48\mathcal{N}(-2,1) + 0.04\mathcal{N}(0,16) + 0.48\mathcal{N}(2,1)$
    \item \textbf{Poorly-Separated (PS)}: A single normal with high overlap with the null, $f_1(z) = \mathcal{N}(0,9)$.
\end{itemize}
Figure~\ref{fig:alternative_densities} shows the densities used in our benchmarks.

For each of the $6$ combinations of the above scenarios, we run $100$ independent trials. Each trial uses $50$ covariates; for all trials with a non-constant prior, $25$ of the variables are used in the true data generating distribution and the other $25$ are null variables with no association with the outcome. To measure sample efficiency, we also vary the sample size from $n=100$ to $n=50K$. The target FDR threshold is set to 10\% for both stages of inference.

We compare BB-FDR to the classic Benjamini-Hochberg (BH) method \cite{benjamini:hochberg:1995:bh}, the recently-proposed NeuralFDR \cite{xia:etal:2017:neuralfdr}, and a fully-Bayesian logistic regression model for $c_i$ in place of the black box prior in \eqref{eqn:bbfdr_generative_model}. For NeuralFDR, we use the default recommended settings, including five random restarts and a ten-layer deep neural network. The fully-Bayesian method uses a standard normal prior on the weights and an inverse-Wishart prior on the variance, with weak hyperpriors. In the nonlinear scenario, we specify all possible pairwise interactions as the covariate set for the fully-Bayesian model to ensure it is well-specified. We fit the model using Polya-gamma sampling \cite{polson:etal:2013:polyagamma} with $5000$ burn-in iterations and $1000$ samples.  For BB-FDR, we use a $50\times200\times200\times2$ network with ReLU activation; for training we use RMS-prop \cite{tieleman:hinton:2012:rmsprop} with dropout, learning rate $3\times10^{-4}$, and batch size $100$, and run for $50$ epochs, with 3 folds to create 3 separate models as in NeuralFDR; we set the $\lambda$ regularization term to $10^{-4}$.

\subsection{Stage 1 performance}
\label{subsec:benchmarks:stage_one}

Figure~\ref{fig:benchmarks} shows the results for the Stage 1 benchmarks, where the goal is to determine for which experiments the treatment had an effect. The four methods generally conserve FDR at the specified $10\%$ threshold, though NeuralFDR seems to systematically violate FDR in the low-sample regime.

Across all experiments, we see that both BH and NeuralFDR under-perform the two Bayesian methods. In the case of BH, this is straight-forward as it uses only the p-value from each experiment and has no notion of side information. NeuralFDR, on the other hand, uses a deep neural network and several random restarts. There are a few possible reasons for its poor performance. First, the NeuralFDR method was reported to be very difficult to train by the original authors, so it is possible that it is simply not finding good fits of the model. Second, BB-FDR assumes that the alternative distribution is conditionally independent of the prior; NeuralFDR makes no such assumption and may lose some power as a result. Finally, NeuralFDR was originally tested on 1- and 2-dimensional problems against relatively weak baselines. Our benchmarks examine its performance in a higher-dimensional setting and with several uninformative features that may make fitting NeuralFDR difficult.

Since the fully-Bayesian method is well-specified in every benchmark, it serves as an oracle model to establish a reasonable upper bound on Stage 1 performance. However, the oracle power depends on the MCMC approximation of the posterior being well-mixed. As the sample size grows, the empirical-Bayes model used by BB-FDR gains an increasingly precise approximation to the true posterior. In the large-sample regime with a well-separated alternative, BB-FDR outperforms even the oracle. Furthermore, the fully-Bayesian method takes several hours to fit in the nonlinear scenarios; BB-FDR fits within a few minutes and can easily be run on a laptop.

\subsection{Stage 2 performance}
\label{subsec:benchmarks:stage_two}
Neither BH nor NeuralFDR provide support for detecting important features (Stage 2), so we could not compare against them. For the Bayesian linear regression, we take the $90\%$ posterior credible interval over the coefficient value for each covariate. If the interval does not contain zero, we reject the null hypothesis and report it as a discovered feature; this approach is standard in the Bayesian literature \cite{gelman:etal:2014:bda3}.

Figure~\ref{fig:benchmarks_covariates} presents the results of the variable selection benchmarks for the poorly-separated alternative distribution; results for the well-separated are similar. We omit the constant scenario, since there are no features to discover. In the small sample regime, the conditional distributions are poor estimators of the conditional null distribution for each feature. This leads to BB-FDR overestimating the number of signal features and violating the FDR threshold. As the sample size grows, the conditional null and black box prior become more accurate, leading to FDR control and higher power, respectively; in the large-sample regime, BB-FDR outperforms the fully-Bayesian approach.

We conclude by noting that BB-FDR is competitive with the fully-Bayesian approach even when the latter is well-specified. In practical data analysis scenarios, such as the cancer study we discuss next, we do not know the true prior function. It may easily contain many higher-order interaction terms that are prohibitive to consider explicitly in a fully-Bayesian model, making BB-FDR a pragmatic choice for real-world scientific datasets.


\section{Cancer drug screening}
\label{sec:cancer}
As a case study of how BB-FDR is useful in practice, we apply it to two high-throughput cancer drug screening studies (Lapatinib and Nutlin-3) from the Genomics of Drug Sensitivity in Cancer (GDSC) \cite{yang:etal:2012:gdsc}. For both drug studies, BB-FDR increases the number of Stage 1 discoveries over classical BH correction; results on NeuralFDR were similar to BH and are omitted. In Stage 2, BB-FDR discovers biologically-plausible genes that may have a causal link to drug sensitivity and resistance. Experimental details are available in the supplement.

\subsection{Analysis overview}
Analysis of the two drug studies broadly follows the two stages outlined in the motivating example in Section~\ref{sec:introduction}. The Stage 1 task is to determine, for a specific drug being tested on a specific cell line, whether the drug had any effect. As with any biological process, natural variation injects randomness at many levels of the experiment: how fast the cells grow, how each cell responds, etc. Thus Stage 1 requires performing statistical hypothesis testing to determine if the cell population after treatment is truly smaller than would be expected from a control (null) population.

The inferential goal in Stage 2 is to gain scientific insight about which genes may be driving drug response. This involves building a statistical model of the relationship between the genomic profile of a cell line and its corresponding treatment response, then performing variable selection on the model. The selected genes form the basis for potential mechanisms of action and future experiments can be designed to test for a causal link or to investigate new drugs that better-target the proteins encoded by the genes.

\subsection{Results}
\label{subsec:cancer:results}
\begin{figure}[th]
\centering
\subfigure[BH on Lapatinib \newline \text{  }(117 discoveries)]{\includegraphics[width=0.21\textwidth]{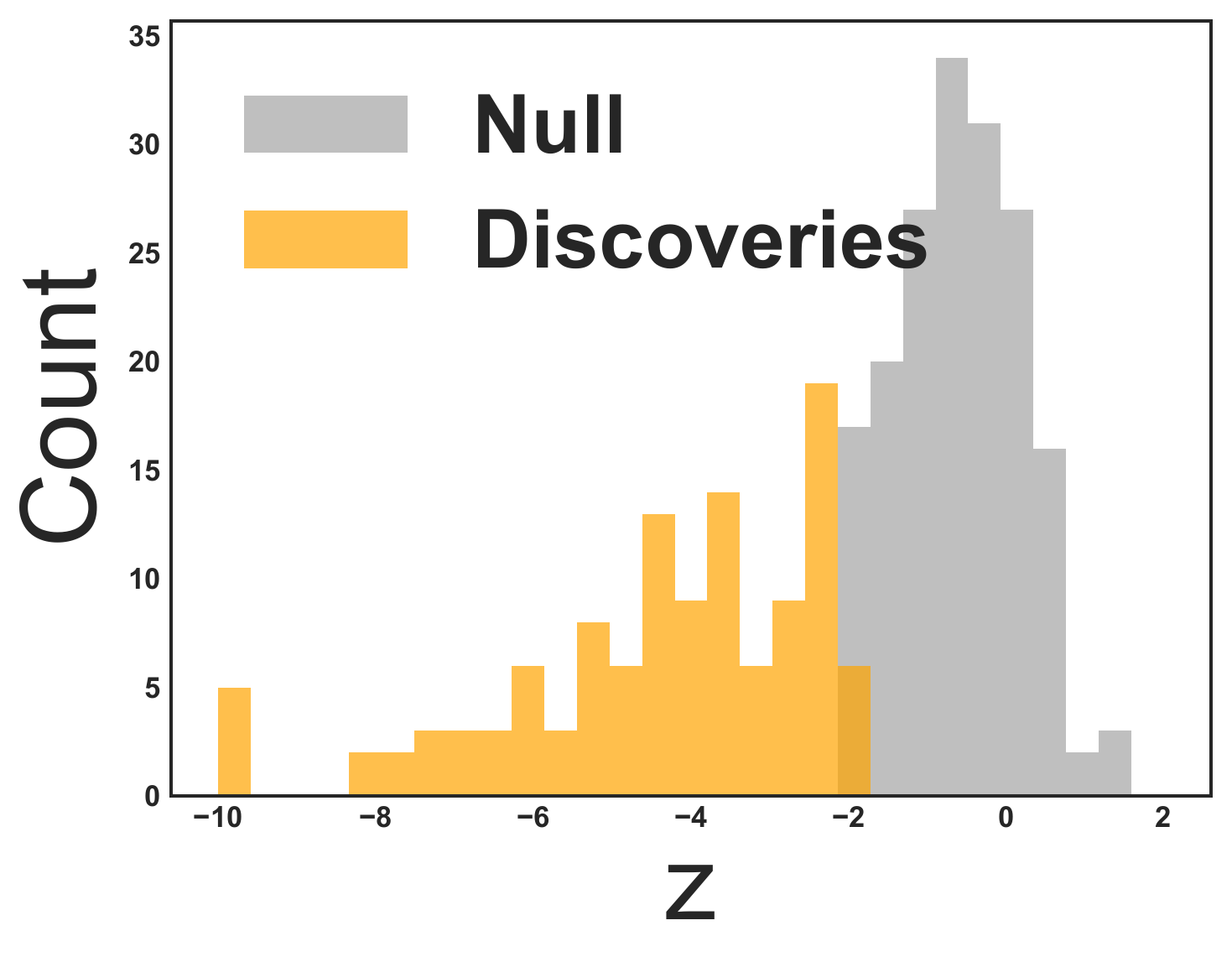}}
\subfigure[BH on Nutlin-3 \newline (151 discoveries)]{\includegraphics[width=0.21\textwidth]{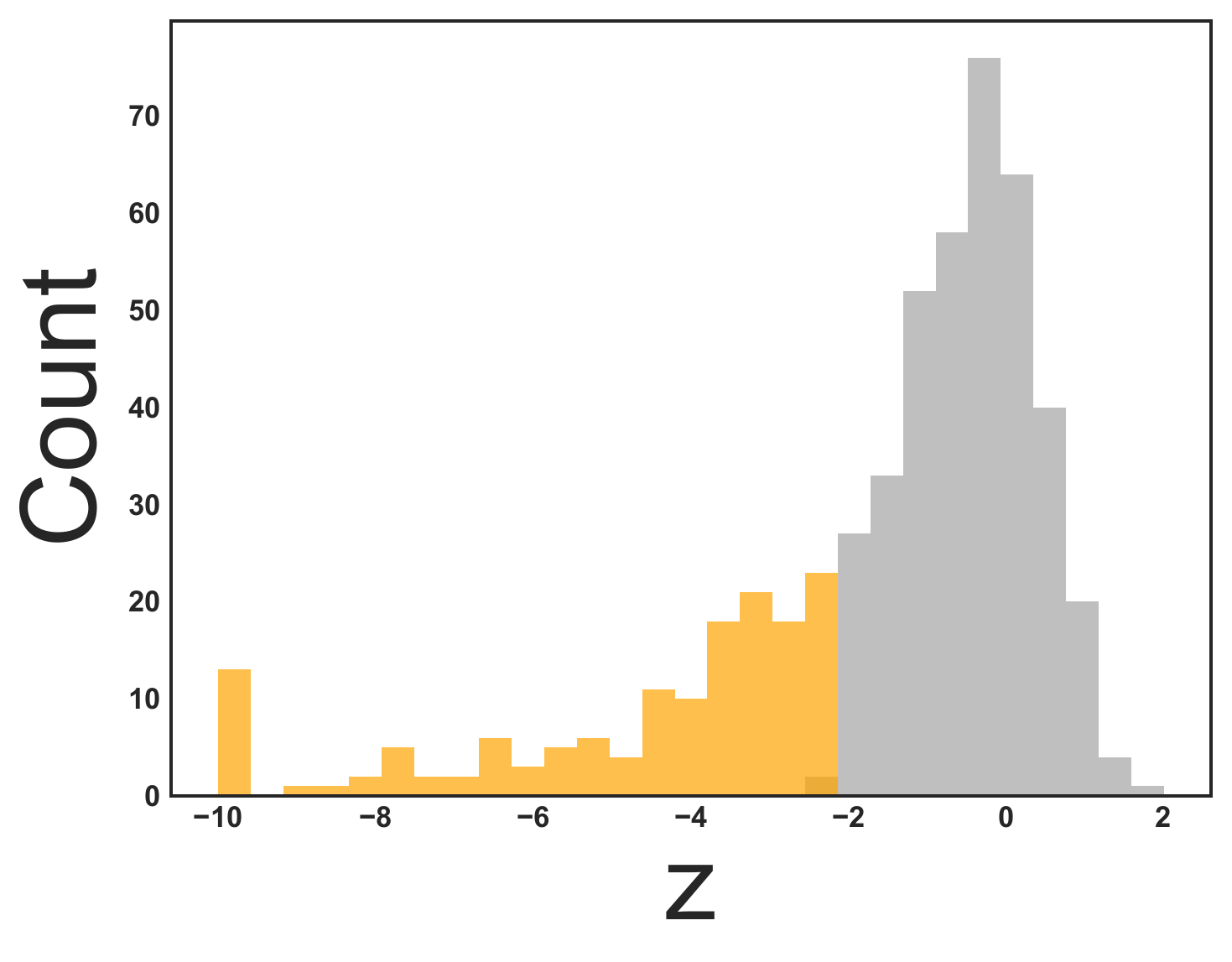}}
\subfigure[BB-FDR on Lapatinib \newline (181 discoveries)]{\includegraphics[width=0.21\textwidth]{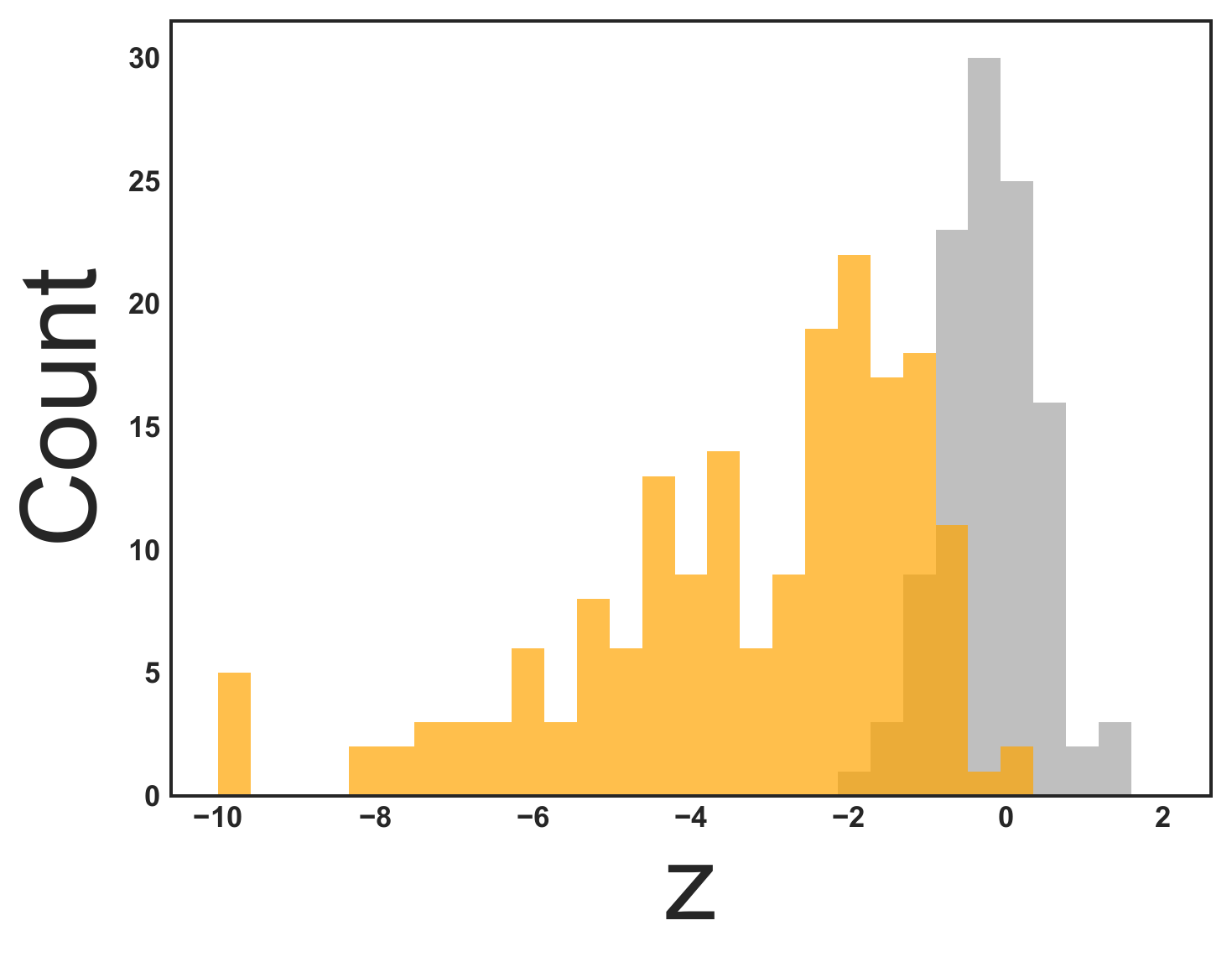}}  
\subfigure[BB-FDR on Nutlin-3 \newline (222 discoveries)]{\includegraphics[width=0.21\textwidth]{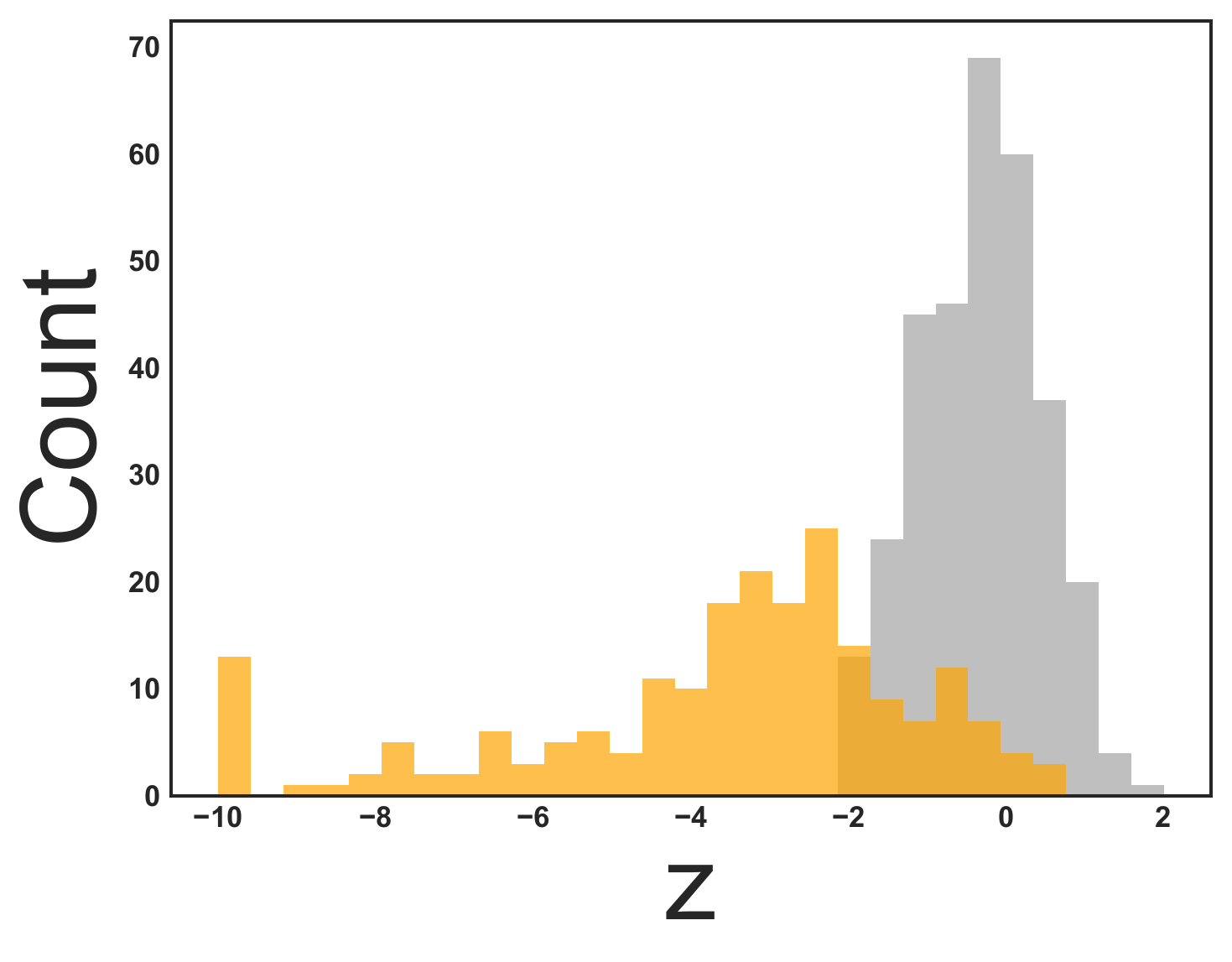}}
\caption[Z-scores for the cancer drug case study]{\label{fig:cancer_discoveries} Discoveries found by BB-FDR on the two drugs, compared to the discoveries found by a naive BH approach. BB-FDR leverages the genomic profiling information of the cell lines to identify $\approx50\%$ more discoveries at the same $10\%$ FDR threshold.}
\end{figure}

Figure~\ref{fig:cancer_discoveries} shows the aggregate number of treatment effects discovered by both BH and BB-FDR. For both drugs, BB-FDR provides approximately a $50\%$ increase in Stage 1 discoveries compared to BH. The genomic profiles of the cell lines provide enough prior information that even some outcomes with a $z$-score above zero are still found to be significant. This flexibility is impossible with classical Stage 1 testing methods like BH that do not consider covariate information.

Table~\ref{tab:cancer_genes} lists the genes reported by BB-FDR in Stage 2. Interpreting the quality of the results requires familiarity with genomics and cancer biology. Below, we briefly detail the scientific rational behind the biological plausibility of the Stage 2 results and refer the reader to \citet{weinberg:2013:biology_of_cancer} for a full review.


\begin{table}
\centering
\begin{tabular}{ll}
\toprule
Lapatinib &  Nutlin-3 \\
\midrule
BRCA1, BRCA2, CDK4 & P300, FLCN, FLT3\\
FGFR2, KIT, MSH2 & MET, KIT, MSH6\\
& SETD2, TP53, BCR-ABL\\
\bottomrule
\end{tabular}
\caption{\label{tab:cancer_genes}Significant gene mutations identified by BB-FDR that are associated with sensitivity and resistance to each drug. Both lists align well with known genomic targets of Lapatinib and Nutlin-3.}
\end{table}

Lapatinib has been approved for the treatment of HER2-positive breast cancers. BB-FDR indicates that BRCA1 and BRCA2 are associated with responses to Lapatinib. Both are tumor suppressor genes that are seen mutated in more than $10\%$ of breast cancers (BRCA stands for ``breast cancer'') and thus cancer type may represent a latent confounder for drug efficacy that induces a conditional dependence. Lapatinib targets over-expression of the gene ERBB2 which can be caused by a mutant CDK4 gene. FGFR2 and KIT are also commonly associated with breast cancers \cite{slattery:etal:2013,zhu:etal:2014} and BRCA1 is known to induce inactivation of the tumor suppressor MSH2 \cite{atalay:etal:2002}. Given Lapatinib's success as a breast cancer drug, the connection between all of the selected genes and breast cancer is a reassuring sign that BB-FDR selected biologically plausible features.

Nutlin-3 is an inhibitor of the oncogene MDM2, which negatively-regulates TP53. When highly over-expressed, MDM2 can functionally inactivate TP53. By targeting MDM2, Nutlin-3 enables a non-mutated (``wild type'') TP53 to trigger apoptosis in cancer cells. However, if TP53 is mutated, Nutlin-3 will be ineffective and hence its mutation state is an important driver of Nutlin-3 sensitivity. When wild type TP53 is present, MET controls the fate of the cell \cite{sullivan:etal:2012}, SETD2 functionality is required to activate TP53 \cite{carvalho:etal:2014:setd2}, P300 mediates TP53 acetylation \cite{reed:quelle:2014:p53}, and BCR-ABL is a gene fusion that induces loss of TP53 \cite{pierce:etal:2000:ablbcr}. These genes interact in complex, non-linear ways, yet BB-FDR is still able to identify them as important. Finally, FLCN is a tumor suppressor gene that can delay cell cycle like TP53 \cite{laviolette:etal:2013}. The mechanism by which FLCN and TP53 are interrelated is currently unclear, representing a potential target for future experiments.

\section{Conclusion}
\label{sec:conclusion}
We presented Black Box FDR (BB-FDR), an empirical-Bayes method that increases statistical power in multi-experiment scientific studies when side information is available for each experiment. BB-FDR combines deep probabilistic modeling with recent multiple testing techniques to boost testing power without sacrificing interpretability. Benchmarks show that BB-FDR outperforms state-of-the-art techniques, often substantially and under a wide array of experimental conditions. BB-FDR also finds more experimental discoveries two cancer drug screening datasets and provides scientific insight into the mechanisms associated with differential treatment response in cancer.

This work was supported by a pilot grant from Columbia University, NIH U54 CA193313, ONR N00014-15-1-2209, ONR 133691-5102004, NIH 5100481-5500001084, NSF CCF-1740833, the Alfred P. Sloan Foundation, the John Simon Guggenheim Foundation, Facebook, Amazon, and IBM.

\bibliography{main}
\bibliographystyle{icml2018}

\clearpage
\appendix

\section{Additional Stage 2 details and results}
For BB-FDR, we fit an XGBoost \cite{chen:guestrin:2016:xgboost} model for each of the three folds and each of the $50$ covariates. We used the XGBoost holdout predictions and the corresponding BB-FDR fold-specific model to calculate null posterior entropies. Each covariate $p$-value was estimated based on $200$ IID Monte Carlo draws and Benjamini-Hochberg correction was applied to each set of $50$ $p$-values at a $10\%$ FDR threshold.

As a practical matter, we recommend checking that estimating the global prior using the standard two-groups model \cite{efron:2008:twogroups} provides a substantially lower number of discoveries than BB-FDR. If it does not, then there is no clear signal available in the covariates. If there is no clear benefit to the added machinery of the black box modeling, then the standard two-groups model should be sufficient.

\begin{figure*}[ptbh]
\centering
\subfigure[Linear (WS)]{\includegraphics[width=0.45\textwidth]{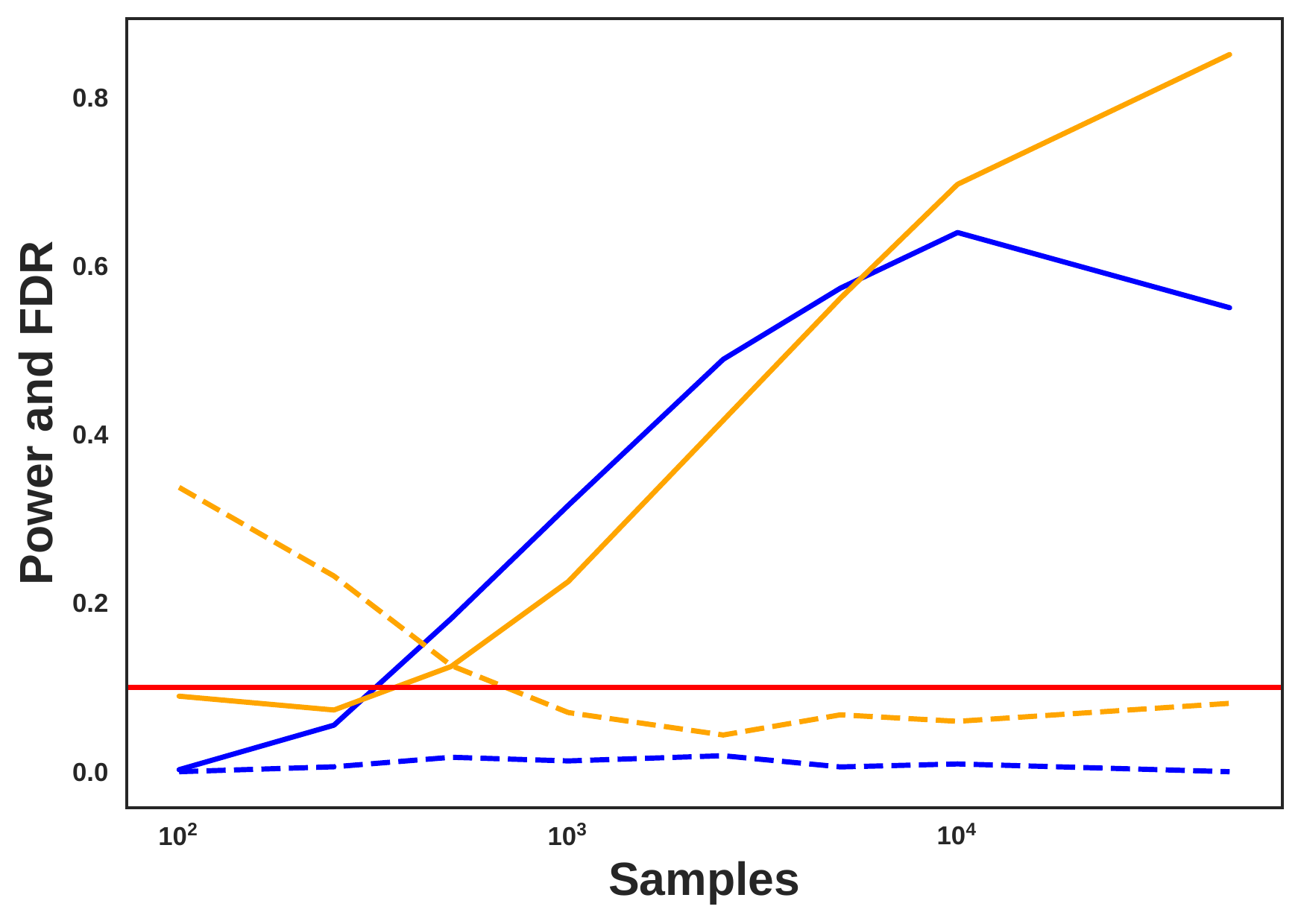}}
\subfigure[Nonlinear (WS)]{\includegraphics[width=0.45\textwidth]{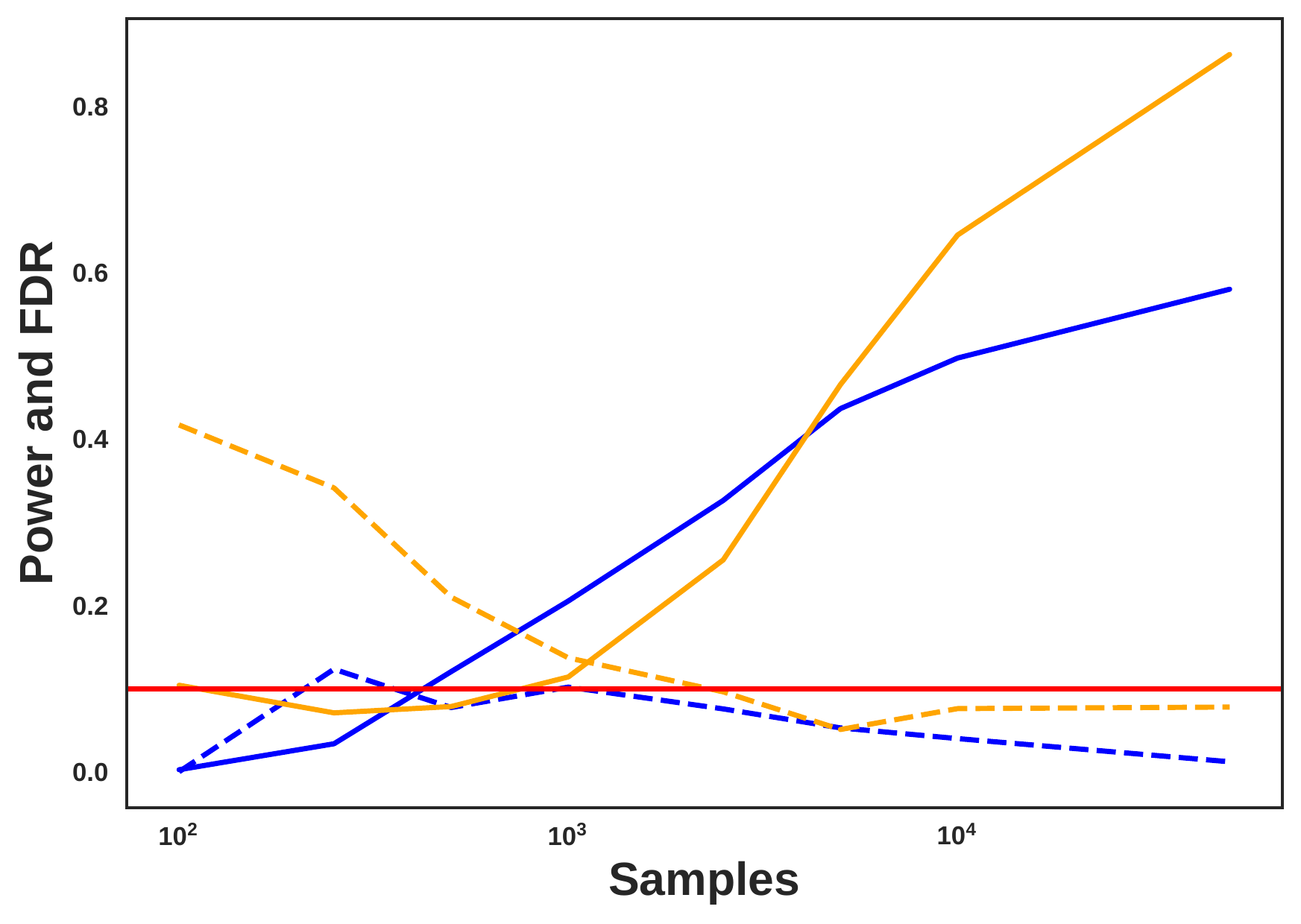}}
\caption{\label{fig:benchmarks_covariates_ws} Variable selection results at a 10\% FDR threshold. Results are similar to that of the poorly-separated alternative in Section 4 of the main text.}
\end{figure*}

Figure~\ref{fig:benchmarks_covariates_ws} shows the analogous results for the Stage 2 benchmarks on the well-separated alternative distribution.

\section{Experimental details for cancer drug screening case study}
We select two drugs in the GDSC dataset with known biological targets: Lapatinib, an EGFR-HERR2 inhibitor; and Nutlin-3, an MDM2 inhibitor. The genomic targets of both drugs are well understood, making them useful drugs on which to validate BB-FDR. Each drug was tested on a large number of cancer cell lines (294 for Lapatinib and 528 for Nutlin-3) to determine treatment efficacy. The results were measured by immunofluorescence; positive (untreated) and negative (unpopulated) controls were decorrelated and temporal batch effects were removed. Positive controls are populations of cells which were never treated and thus their measurements serve as a null distribution. We take the maximum concentration dosage for both drugs and calculate $z$-scores for the treatment response, yielding a series of test statistics. 

Each cell line has also been profiled via WES and we followed \cite{iorio:etal:2016:gdsc_ccle} in filtering down to a set of relevant, protein-changing gene mutations. We throw out any cell lines that were not sequenced and any genes whose mutation state was constant across all cell lines. For each cell line, we then get a binary feature vector representing the genomic profile of the experiment (66 features for Lapatinib and 71 for Nutlin-3). These features can then serve as an important source of prior information for the sensitivity or resistance of each cell line to the given drug.

A standard approach is to build a linear model of the genomic features to predict the outcome measurement. However, cellular processes are driven by complex, non-linear interactions known as gene regulatory networks \cite{karlebach:Shamir:2008:nonlinear_gene_regulation}. A linear model of the interaction of genes is unlikely to be well-specified and will likely be underpowered, as in the non-linear prior scenario from Section 4.3. BB-FDR is a natural fit given its flexibility to model highly non-linear covariate dependencies.

We kept the majority of the BB-FDR settings the same as in Section 4. However, to account for the smaller sample sizes here, we changed the batch size to 10 and created 10 folds instead of 3 to maximize training data available to each model.

For all of our results, we note that the Stage 2 discoveries may be confounded by cancer cell type and do not necessarily represent \textit{causal} links. For instance, BRCA1 and BRCA2 are frequently found mutated in breast cancer cells.

\end{document}